\documentclass[pdflatex,sn-mathphys-num]{sn-jnl}
\usepackage{graphicx}%
\usepackage{multirow}%
\usepackage{amsmath,amssymb,amsfonts}%
\usepackage{amsthm}%
\usepackage{mathrsfs}%
\usepackage[title]{appendix}%
\usepackage{xcolor}%
\usepackage{textcomp}%
\usepackage{manyfoot}%
\usepackage{booktabs}%
\usepackage{algorithm}%
\usepackage{algorithmicx}%
\usepackage{algpseudocode}%
\usepackage{listings}%
\theoremstyle{thmstyleone}%
\usepackage{float}
\usepackage{mathrsfs}
\theoremstyle{thmstyletwo}%

\theoremstyle{thmstylethree}%

\begin{document}

\title[TwinOR]{TwinOR: Photorealistic Digital Twins of Dynamic Operating Rooms for Embodied AI Research}
\author*[1]{Han Zhang} \email{hzhan206@jhu.edu}
\author[1]{Yiqing Shen}
\author[1]{Roger D. Soberanis-Mukul} 
\author[1]{Ankita Ghosh}
\author[1]{Hao Ding} 
\author[1]{Lalithkumar Seenivasan}
\author[1,2]{Jose L. Porras} 
\author[1]{Zhekai Mao} 
\author[1]{Chenjia Li} 
\author[1]{Wenjie Xiao}  
\author[2]{Lonny Yarmus}
\author[2]{Angela Christine Argento} 
\author[2]{Masaru Ishii}
\author*[1]{Mathias Unberath} \email{unberath@jhu.edu}
\affil[1]{\orgname{Johns Hopkins University}, \city{Baltimore}, \state{MD}, \country{USA}}
\affil[2]{\orgname{Johns Hopkins Medical Institutions}, \city{Baltimore}, \state{MD}, \country{USA}}

\abstract{
\textbf{Purpose:} 
Developing embodied AI for intelligent surgical systems requires safe, controllable environments for continual learning and evaluation. However, safety regulations and operational constraints in operating rooms (ORs) limit agents from freely perceiving and interacting in realistic settings. Digital twins provide high-fidelity, risk-free environments for exploration and training. How we may create photorealistic and dynamic digital representations of ORs that capture relevant spatial, visual, and behavioral complexity remains an open challenge.

\textbf{Methods:} 
We introduce \textsc{TwinOR}, a real-to-sim infrastructure for constructing photorealistic and dynamic digital twins of ORs for embodied AI research. The system reconstructs static geometry from pre-scan videos and continuously models human and equipment motion through multi-view perception of OR activities. The static and dynamic components are fused into an immersive 3D environment that supports controllable simulation and facilitates future embodied exploration.

\textbf{Results:}
The proposed framework reconstructs complete OR geometry with centimeter-level accuracy while preserving dynamic interaction across surgical workflows, enabling realistic renderings and a virtual playground for embodied perception benchmarks. In our experiments, \textsc{TwinOR} synthesizes stereo and monocular RGB streams as well as depth observations for geometry understanding and visual localization tasks. Models such as FoundationStereo and ORB-SLAM3 evaluated on \textsc{TwinOR}-synthesized data achieve performance within their reported accuracy ranges on real-world indoor datasets, demonstrating that \textsc{TwinOR} provides sensor-level realism sufficient for emulating real-world perception and localization challenges in dynamic OR scenes.

\textbf{Conclusion:} 
By establishing a perception-grounded real-to-sim pipeline, \textsc{TwinOR} enables the automatic construction of dynamic, photorealistic digital twins of ORs. As a safe and scalable environment for experimentation and benchmarking, \textsc{TwinOR} opens new opportunities for translating embodied intelligence from simulation to real-world clinical environments, and sets the stage for future research on interaction, autonomy, and human–robot collaboration in the OR.
}

\maketitle

\section{Introduction}\label{sec1}
Operating rooms (ORs) are among the most complex and safety-critical environments in healthcare, where every action requires precise coordination among surgeons, nurses, and increasingly, intelligent robotic systems~\cite{ozsoy_oracle_2024,killeen_fluorosam_2026,zhang_straighttrack_2024,zhang_did_2025}. Despite continuous advances in surgical robotics systems and Artificial Intelligence (AI) technologies, modern surgical practice still faces persistent challenges: a global shortage of skilled medical staff, rising operational costs, and the ever-present risk of human error. These challenges highlight the urgent need for intelligent automation that can augment human expertise and perform surgical-support tasks reliably and safely. 

Recent progress in embodied AI, which integrates perception, reasoning, and action to enable agents to interact with physical environments, offers a compelling opportunity to introduce adaptive intelligence into surgical settings~\cite{liu_vision_2025}. Embodied agents capable of learning through interaction could support context-aware assistance, instrument handovers, or autonomous navigation in the OR~\cite{kumar_health_2024}. However, developing such systems requires large volumes of realistic multi-modal data and safe, controllable environments for continual learning and evaluation. Collecting these data and testing embodied AI systems directly in live surgeries are often impractical due to safety constraints, limited accessibility, and the risk of disrupting clinical operations. As a result, researchers increasingly rely on synthetic data generation or simulated environments~\cite{tagliabue_soft_2020,11223235}. Nevertheless, existing synthetic approaches often struggle to capture the full spatiotemporal complexity of real ORs, leading to unrealistic visual appearance or temporally inconsistent motion~\cite{ozsoy_oracle_2024}.

Digital twin technologies provide high-fidelity virtual replicas of physical systems that remain synchronized with the real world~\cite{ding_digital_2024}. Complementary to powerful simulation frameworks, digital twins have proven effective in domains such as manufacturing, autonomous driving, and smart cities, where they serve as platforms for reinforcement learning, predictive control, and synthetic data generation~\cite{oo_digital_2025,liu_dart_2025,killeen_silico_2023,munawar_virtual_2022}. In surgical contexts, digital twins have been explored for training, visualization, and robotic skill transfer, demonstrating their potential to improve learning and safety without disrupting clinical practice~\cite{killeen_stand_2024,ding_towards_2024,ding_towards_2025}. However, most existing OR-oriented digital twin efforts remain limited to 2D abstractions or quasi-static representations~\cite{perez_privacy-preserving_2025,shen_online_2025}, with only a few recent proof-of-concept systems demonstrating partial dynamic reconstruction~\cite{hein_creating_2024,kleinbeck_neural_2024}. To our knowledge, current OR digital twin approaches do not provide an efficient means of constructing dynamic digital twins that capture accurate geometry, appearance, and continuously evolving human and equipment motion. Such capabilities are critical for systems that require temporally coherent and physically grounded interactions. As a result, constructing realistic and dynamic OR environments remains time-consuming and largely unexplored, presenting a significant bottleneck for embodied AI research in safety-critical clinical settings.

To address this challenge, we introduce \textsc{TwinOR}, a perception-grounded real-to-sim infrastructure for constructing photorealistic and dynamic digital twins of ORs. \textsc{TwinOR} reconstructs static elements of the OR and key equipment from pre-scan videos and continuously models human and equipment motion using automated multi-view perception. These static and dynamic components are fused into a high-fidelity 3D environment that supports immersive viewpoint synthesis, dynamic replay, and controllable simulation. In conjunction with existing simulation frameworks, \textsc{TwinOR} enables the automatic creation of sensor-realistic OR environments grounded in real clinical data.

We evaluate \textsc{TwinOR} across both simulated and real surgical workflows captured in ORs, assessing geometric reconstruction accuracy, photometric realism, dynamic perception accuracy, and its utility as an enabling infrastructure for embodied AI research. Experimental results demonstrate that TwinOR achieves centimeter-level, room-scale geometric accuracy and photorealistic visual fidelity while modeling human and equipment motion observed during surgical procedures. We further validate the feasibility of \textsc{TwinOR} for embodied perception and localization tasks through stereo depth estimation and visual SLAM experiments. Systems evaluated on TwinOR-synthesized data achieve performance comparable to real-world benchmarks, providing sensor-level photometric realism and temporal coherence in dynamic OR scenes.

In summary, the key contribution of this work lies in the development of \textsc{TwinOR}, a photorealistic and dynamic digital twin framework that unifies static geometry reconstruction and continuous motion perception, supporting embodied AI research in realistic, room-scale OR environments.

\begin{figure}[t]
    \centering
    \includegraphics[page=1,width=\textwidth,trim={0 330 0 0}, clip]{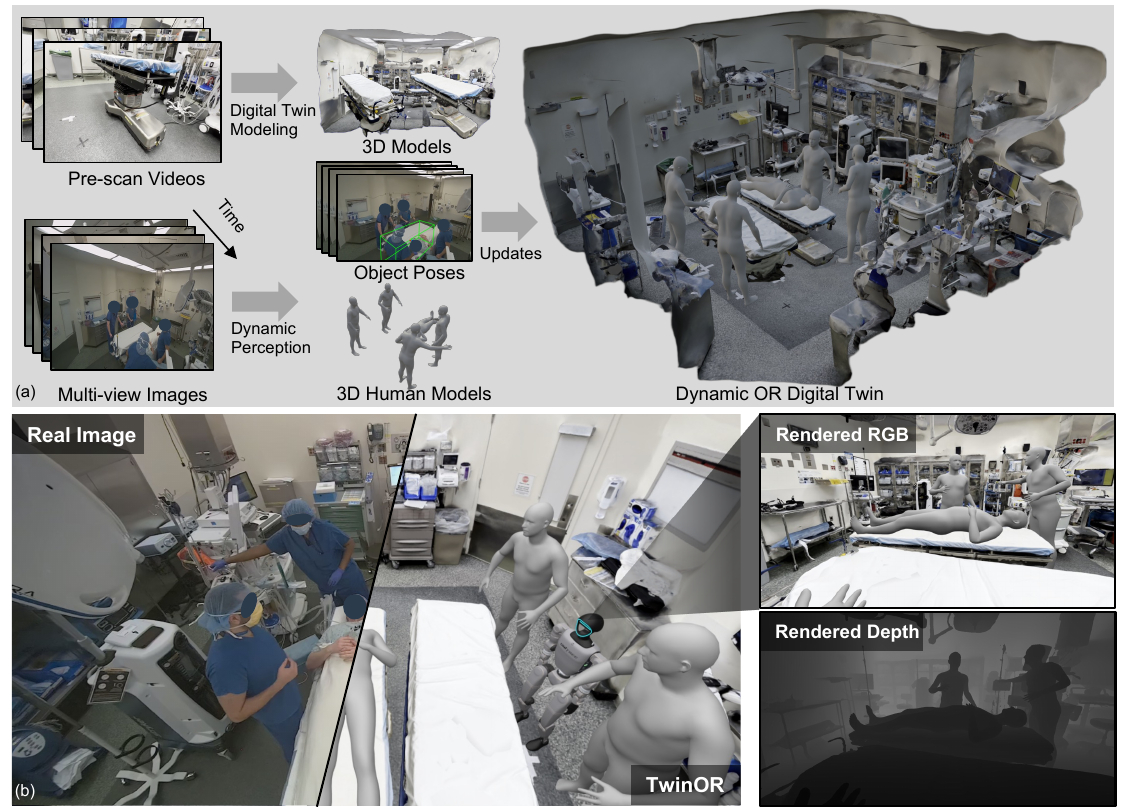}    
    \caption{System overview of the \textsc{TwinOR} framework. (a) \textsc{TwinOR} reconstructs 3D geometry and captures dynamic motion in ORs to create photorealistic digital twins. (b) These digital environments enable controllable synthetic data generation and embodied AI experiments in realistic OR settings.}
    \label{teaser}
\end{figure}

\section{Method}\label{sec2}
\subsection{System Architecture}
The proposed framework integrates high-fidelity 3D reconstruction and multi-view perception to model the dynamic OR. Initially, \textsc{TwinOR} reconstructs the 3D model of the OR and equipment from casually captured pre-scan video. During surgical procedures, multi-view camera system captures and estimates the 3D poses of personnel and equipment. These data streams are continuously fused into a virtual environment implemented in \textsc{Blender}, supporting photorealistic rendering, dynamic replay and, interactive simulation. An overview of the system architecture is shown in Fig.~\ref{teaser}. 

\subsection{Camera Infrastructure}
The multi-view camera system comprises four wall-mounted stereo RGB cameras (ZED-X StereoLabs) positioned near room corners to capture synchronized views. Temporal alignment across all devices was hardware-maintained via Precision Time Protocol, achieving sub-millisecond synchronization. Camera intrinsics were adopted from factory calibration, and inter-camera extrinsics were determined using a custom light wand containing three fixed spherical markers with known inter-marker distances. The wand was moved throughout the OR to cover the full capture volume, providing dense cross-view correspondences for all cameras. Initial relative poses were estimated via a linear PnP solution and subsequently refined through global bundle adjustment. This procedure yielded sub-pixel reprojection accuracy and a metrically consistent, time-synchronized foundation for multi-view 3D perception.

\subsection{Digital Twin Modeling}
\label{sec:assetmodeling}

This phase produces a high-fidelity 3D model, encompassing both static structures (walls, ceiling, fixtures) and movable medical equipment. We follow the reconstruction strategy described in our prior work~\cite{kleinbeck_neural_2024}, employing a customized implementation of \textsc{Neuralangelo}~\cite{li_neuralangelo_2023} to recover dense neural surfaces from handheld pre-scan RGB videos. The resulting models were metrically scaled and aligned to the camera rig via registration with fused 3D point cloud from multi-view stereo cameras. Representative reconstructed models are shown in Fig.~\ref{Reconstruction}.

\subsubsection*{Room Reconstruction}
Each OR was scanned with a $\sim$20-minute wide-angle handheld video recording to ensure full coverage. The reconstructed neural surfaces were post-processed to remove artifacts, fill missing geometry near reflective or textureless regions, and partition meshes into two $\sim$400k faces segments textured at 4K resolution, balancing visual fidelity and runtime performance. Reconstruction required approximately 40–50 hours per room on an NVIDIA Quadro RTX~6000 GPU.

\subsubsection*{Equipment Reconstruction}
Each movable medical equipment item was reconstructed with a $\sim$2-minute wide-angle handheld video with full 360° coverage. In our experiments, we reconstructed four representative equipment: the operating table, the Ion surgical robot, the C-arm X-ray system, and the stretcher. The resulting meshes were simplified to 20k–40k faces and textured at 2K resolution to balance geometric detail and rendering efficiency within the simulation environment.

\subsection{Dynamic Perception}
\label{sec:dynamicperception}
\subsubsection*{3D Human Modeling}

Each individual in the OR is represented using the statistical parametric human mesh model \textsc{SMPL}~\cite{loper_smpl_2023}. We adopt a top-down multi-view approach combining detection, triangulation, and model fitting. A fine-tuned human detector is first applied to all RGB views to obtain person bounding boxes. Within each box, the pre-trained 2D pose estimator \textsc{ViTPose+}~\cite{xu_vitpose_2023} predicts 17 human anatomical keypoints, which are subsequently triangulated across calibrated views to recover 3D joint locations. The triangulated 3D joints are then fitted to the \textsc{SMPL} model for full-body reconstruction.

Temporal association is achieved by matching individuals across frames based on 3D spatial proximity and pose similarity. To mitigate perception failures caused by occlusions or rapid human motion, we further apply multi-view optimization and temporal smoothing. Specifically, a multi-stage optimization then refines the \textsc{SMPL} parameters by jointly fitting to both 2D and 3D keypoints, while enforcing motion priors and temporal smoothness to ensure physically plausible trajectories and temporally coherent motion~\cite{zeng_smoothnet_2022}. The pipeline was implemented and extended based on the EasyMocap~\cite{easymocap} framework.

\subsubsection*{Equipment Motion Modeling}
To model equipment motion in the OR, we estimate their 6-DoF poses from multi-view stereo observations. Each frame is processed using the Segment Anything Model 2 (SAM2)~\cite{ravi_sam_2024} to segment the target equipment across all camera views. With depth maps from the stereo cameras, the segmented regions are back-projected into 3D and fused into an equipment-specific point cloud.

The fused point cloud is first coarsely aligned with the corresponding CAD model through RANSAC-based feature matching, where local geometric descriptors are extracted using Fast Point Feature Histograms (FPFH)~\cite{rusu_fast_2009}. The alignment is further refined using a color Iterative Closest Point (ICP) algorithm~\cite{park_colored_2017}, which jointly optimizes geometric and photometric consistency. The estimated 6-DoF poses are subsequently used to update the equipment motion in the simulation.

\section{Experiments and Results}\label{sec4}
\subsection{Experimental Setup}

We evaluated the \textsc{TwinOR} across surgical workflows. The dataset comprises two simulated surgeries conducted in OR-A, covering the complete clinical workflow from preoperative preparation to patient exit. To assess generalizability, we collected a real patient case in OR-B, which features distinct room geometry, equipment layout, and clinical functionality. Each case includes synchronized multi-view RGB recordings at 1080p and 30~FPS and corresponding pre-scan videos for static scene modeling. The mean reprojection error of multi-camera calibration was $0.44$~px for OR-A and $0.55$~px for OR-B. All reconstruction and perception processes were performed offline to ensure globally consistent geometry and photorealistic rendering. The evaluation focuses on three aspects: (a) digital twin evaluation in terms of static reconstruction and dynamic perception accuracy, (b) geometric perception validation, and (c) visual localization validation for embodied AI tasks.

\subsection{Digital Twin Evaluation}

\begin{figure}[t]
    \centering
    \includegraphics[page=2,width=\textwidth,trim={0 510 0 0}, clip]{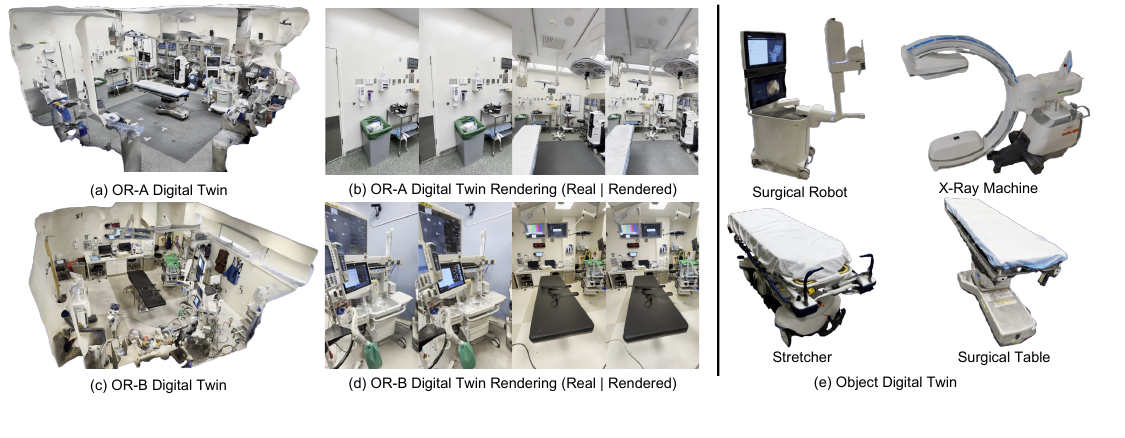}
    \caption{Static reconstruction quality of \textsc{TwinOR}, illustrating photorealistic digital twins of ORs and representative equipment.}
    \label{Reconstruction}
\end{figure}

We first evaluated the geometric and photometric fidelity of \textsc{TwinOR} reconstructions. For static reconstruction, neural reconstruction achieved high visual quality, with mean scores of \textbf{SSIM~0.90/0.92}, \textbf{PSNR~27.7/25.4\,dB}, for OR-A and OR-B, respectively. While the implicit neural representation provides superior visual fidelity, it is computationally expensive for real-time rendering and physics-based interaction. To enable efficient simulation, we converted the neural fields into mesh-based assets via surface extraction and texture baking. The resulting textured meshes maintained the realism with \textbf{SSIM~0.72/0.82}, \textbf{PSNR~18.6/18.8\,dB}. To quantify geometric accuracy, we compared our reconstructions against sparse point clouds generated by COLMAP~\cite{schonberger_structure--motion_2016}, a widely adopted state-of-the-art structure-from-motion framework commonly used in 3D reconstruction evaluations~\cite{schops_multi-view_2017,knapitsch_tanks_2017}. Both the neural surface reconstructions and COLMAP point clouds were aligned to stereo RGB point clouds captured by our calibrated camera system to ensure metric scale consistency. Under this alignment, \textsc{TwinOR} achieved mean chamfer distances of \textbf{14.1\,mm} (OR-A) and \textbf{22.7\,mm} (OR-B), corresponding to centimeter-level spatial accuracy. Representative qualitative results for room and equipment reconstruction are shown in Fig.~\ref{Reconstruction}.

\begin{figure}[t]
    \centering
    \includegraphics[page=3,width=\textwidth,trim={0 370 0 0}, clip]{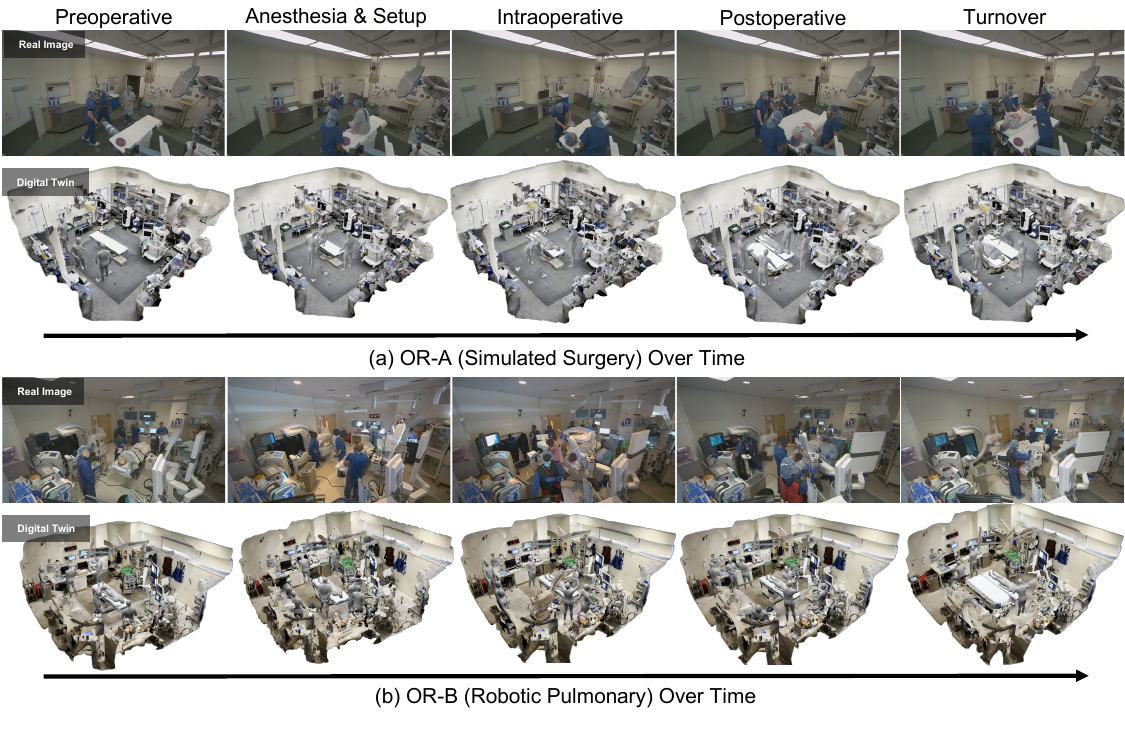}
    \caption{Dynamic perception fidelity of \textsc{TwinOR}. Reconstructed motion sequences of simulated (a) and real (b) surgical workflows. \textsc{TwinOR} reproduces human and equipment motion with temporal coherence and visual realism.}
    \label{Perception}
\end{figure}

For 3D human estimation evaluation, we uniformly sampled 100 scenes across five representative procedural phases from two simulated full surgical workflows in OR-A. Ground-truth skeletons were obtained by triangulating manually annotated 2D keypoints across views. All metrics were computed on visible joints only, as our goal is to assess the spatial accuracy of detected human poses rather than inferring occluded joints. \textsc{TwinOR} achieved a mean 3D Percentage of Correct Parts (PCP3D@0.5) of \textbf{98.34\%}, and a Mean Per Joint Position Errors (MPJPE) of \textbf{3.52\,cm
}. For equipment motion estimation, we evaluated three representative movable items, including the operating table, the C-arm imaging system, and the patient stretcher, which capture the major rigid-body motions typically observed in OR. Smaller tools and deformable items were excluded due to limited observability. The object pose evaluation was conducted on 10 randomly sampled scenes across different phases and ORs with fully or partially visible instances of each specific item. Ground-truth 6-DoF poses were obtained by manually aligning equipment CAD models with fused stereo RGB point clouds. \textsc{TwinOR} achieved mean translational and rotational errors of \textbf{9.12\,cm} and \textbf{4.57°}, respectively, with a mean 3D bounding box Intersection-over-Union (IoU) of \textbf{0.82}. These quantitative results demonstrate that \textsc{TwinOR} is able to capture both human and equipment motion at the room scale in real-world OR environments. Qualitative results across representative procedural phases are shown in Fig.~\ref{Perception}, illustrating the temporal evolution of the OR digital twin. 

In addition to the main experiments, we qualitatively demonstrated \textsc{TwinOR} on a real patient robotic pulmonology procedure, which represents a more complex interventional OR setting with a crowded clinical team, large robotic systems, and mobile imaging equipment. Representative visualizations from this case are shown in Fig.~\ref{Perception}(b), highlighting the applicability of \textsc{TwinOR} in high-complexity real scenarios.

Overall, \textsc{TwinOR} demonstrates the ability to maintain spatial accuracy and temporal coherence throughout dynamic surgical scenes, providing a geometrically and photometrically consistent digital twin foundation for OR environments.

\subsection{Geometric Perception for Embodied AI}
\begin{figure}[t]
    \centering
    \includegraphics[page=4,width=\textwidth,trim={0 530 0 0}, clip]{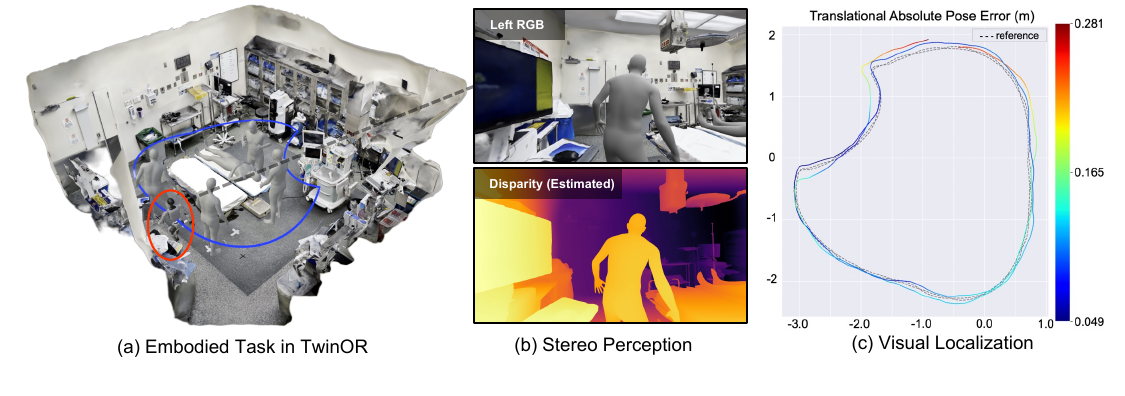}
    \caption{Validation of embodied perception and localization tasks in the \textsc{TwinOR}.}
    \label{Embodied_Tasks}
\end{figure}
\begin{table}[h]
\centering
\small
\begin{tabular}{lcccccc}
\toprule
 & \multicolumn{3}{c}{\textbf{Disparity-domain}} 
 & \multicolumn{3}{c}{\textbf{Depth-domain}} \\
\cmidrule(lr){2-4} \cmidrule(lr){5-7}
 & EPE  & Bad-2px  & D1-all 
 & MAE  & AbsRel  \\
\midrule
\textsc{TwinOR} &
0.378\,px & 1.5\% & 0.6\% &
7\,mm & 0.5\%  \\
\bottomrule
\end{tabular}
\caption{Stereo-depth evaluation on \textsc{TwinOR}.}
\label{tab:stereo_depth_eval_overall}
\end{table}

Accurate geometric perception is fundamental for embodied agents to reason about spatial structures, perform manipulation, and plan interactions. To evaluate whether \textsc{TwinOR} provides sensor-realistic geometry for perception tasks, we conducted zero-shot stereo depth estimation experiments using \textsc{FoundationStereo}~\cite{wen_foundationstereo_2025}, a state-of-the-art foundation model for stereo matching. A virtual stereo camera (1080p resolution, 8\,cm baseline) was placed along a circular trajectory around the operating table in the dynamic OR-A environment, generating 800 stereo pairs with ground-truth disparity and depth from the simulator. 

As shown in Table~\ref{tab:stereo_depth_eval_overall} and Fig.~\ref{Embodied_Tasks}(b) , the model achieved low errors in the disparity domain, with an average per-pixel disparity error (EPE) of \textbf{0.378\,px}, only \textbf{1.5\%} of pixels having disparity error $>$2\,px (Bad-2px), and \textbf{0.6\%} exceeding both 3\,px and 5\% of the ground-truth disparity (D1-all), closely matching the performance range reported on real indoor benchmarks~\cite{wen_foundationstereo_2025}. After converting the predicted disparities into metric depth using the known camera intrinsics and baseline, the model achieved a mean absolute error (MAE) of \textbf{7\,mm}, absolute relative error (AbsRel) of \textbf{0.5\%}, demonstrating sub-centimeter geometric accuracy and strong photometric consistency.

These findings indicate that real-world vision foundation models generalize effectively to the \textsc{TwinOR} environment without adaptation. The alignment between disparity and depth-domain performance confirms that \textsc{TwinOR} preserves sensor-level realism, enabling embodied agents to perform reliable stereo reconstruction, metric depth estimation, and 3D spatial reasoning within realistic surgical scenes.

\subsection{Visual Localization for Embodied AI}

To evaluate \textsc{TwinOR}’s spatial–perceptual realism, we conducted a visual localization and trajectory estimation experiment using a simulated Unitree G1 humanoid robot within the dynamic digital twin of the OR-A. As illustrated in Fig.~\ref{Embodied_Tasks}(a), the simulated robot followed a predefined circular trajectory (blue) around the operating table, covering a total path length of approximately 27\,m. In this experiment, an RGB-D camera is simulated by synthesizing RGB images and depth maps from the robot’s head at 1080p and 30~fps, while ground-truth 6-DoF poses were exported from Blender for quantitative evaluation.

The rendered sequences were processed using ORB-SLAM3~\cite{campos_orb-slam3_2021} in RGB-D mode to recover the camera trajectory purely from visual input. As shown in Fig.~\ref{Embodied_Tasks}(c), the estimated trajectory closely aligned with the ground truth, achieving an Absolute Trajectory Error (ATE) of \textbf{0.12\,m} and a Relative Pose Error (RPE) of \textbf{0.066\,m}. These values fall within the typical error range reported for real-world indoor SLAM benchmarks~\cite{vedadi_comparative_2023}, indicating that visual localization performance in \textsc{TwinOR} closely matches that observed in physical environments. This confirms that \textsc{TwinOR} provides photometrically and geometrically consistent visual observations suitable for reliable localization and mapping.

Overall, these results demonstrate \textsc{TwinOR}’s capability for embodied visual simulation, showing that agents can perceive and localize using realistic sensor observations with coherent spatial and temporal cues. Even under dynamic conditions with independently moving staff and equipment, the robot maintained stable localization, validating \textsc{TwinOR} as a reliable environment for benchmarking embodied AI algorithms in surgical scene understanding and spatial reasoning.

\section{Discussion}\label{sec6}

Our experiments demonstrate that \textsc{TwinOR} achieves high geometric and photometric fidelity, ensuring spatial-perceptual coherence in real OR scenes. The accurate geometric perception and visual localization demonstrate that \textsc{TwinOR} effectively reduces the real-to-sim gap between real and synthetic OR environments. By unifying 3D reconstruction with multi-view dynamic perception, \textsc{TwinOR} establishes a perception-grounded digital twin that reflects real surgical workflows. In this role, \textsc{TwinOR} complements existing simulation frameworks by providing a scalable and cost-efficient environment upon which downstream simulation, benchmarking, and embodied agent development can be built, while minimizing the need for repeated physical trials.

Despite these advantages, several limitations remain. The current system is designed as an offline pipeline with a computationally intensive reconstruction stage. Converting implicit neural reconstructions into mesh-based assets also introduces a trade off between visual fidelity and simulation efficiency. Improving such gap under real-time constraints, such as 3D Gaussian Splatting or hybrid representations, remains an open direction. \textsc{TwinOR} focuses on room-scale geometry and large rigid equipment, and does not model small instruments, articulated tools, or deformable tissues, as all visible assets must be pre-scanned prior to integration. As a result, fine-grained physical interactions such as finger-level manipulation or tool--tissue contact cannot be represented, and the current tracking accuracy remains insufficient for such fine manipulation scenarios. In addition, severe occlusions or rapid motion in ORs may still lead to temporal inconsistencies in human and object pose estimation, which may require explicit occlusion-aware reasoning or longer-horizon modeling. Finally, although \textsc{TwinOR} is designed to be modular and room-agnostic, systematic generalization across a broader range of hospitals, clinical specialties, layouts, and lighting conditions has yet to be evaluated.

Looking forward, \textsc{TwinOR} provides a foundation for future embodied AI research in OR contexts when integrated with downstream simulation and learning frameworks. Future studies will explore the use of \textsc{TwinOR} for higher-level embodied AI tasks, such as navigation, policy learning, and decision-making. Moreover, incorporating real-time updates, deformable object modeling, and additional perception capabilities would further enhance realism and move toward more comprehensive and physically grounded OR digital twins. Beyond geometric and kinematic modeling, integrating higher-level semantic representations remains an important direction for future work. Ultimately, \textsc{TwinOR} represents a stepping stone toward the next generation of intelligent ORs in which robots and clinical staff can collaborate safely and effectively.

\section{Conclusion}\label{sec7}
We presented \textsc{TwinOR}, a framework for creating photorealistic and dynamic OR digital twins. It achieves centimeter-level geometric accuracy and photorealistic visual fidelity, enabling reliable perception and localization tasks in realistic surgical environments. By providing a scalable, perception-grounded real-to-sim infrastructure, \textsc{TwinOR} represents an important step toward translating embodied intelligence from simulation to real-world clinical settings.

\backmatter

\bmhead{Supplementary information}
The video demonstrates our OR digital twin.

\bmhead{Acknowledgment}
We acknowledge support from the National Science Foundation under Award No.~2239077, and Johns Hopkins University Internal Funds.

\bmhead{Ethical approval} 
Data collection procedures were approved under HIRB00016983 and IRB00421946.

\bmhead{Conflict of interest}
The authors have no conflict of interest to declare.

\bibliography{sn-bibliography}
\end{document}